# State Space System Modelling of a Quad Copter UAV

Zaid Tahir
zaid.tahir@york.ac.uk

Waleed Tahir
waleedt@bu.edu

Saad Ali Liaqat
sa.liaqat@gmail.com

*Abstract*

*In this paper, a linear mathematical model for a quad copter unmanned aerial vehicle (UAV) is derived. The three degrees of freedom (3DOF) and six degrees of freedom (6DOF) quad copter state-space models are developed starting from basic Newtonian equations. These state space models are very important to control the quad copter system which is inherently dynamically unstable.*

## 1. Introduction

In the most recent couple of decades fast improvement has been completed in multi-mission proficient, small unmanned aerial vehicles. They are picking up significance because of their capacity to replace manned aerial vehicles in standard and additionally hazardous missions, which thus diminish expenses of numerous aerial operations. These aerial vehicles are being utilized in different nonmilitary missions, for example, news agencies, climate checking and by law authorization offices. Thus they are likewise being widely utilized as a part of military operations, for example, insight information gathering, observation and surveillance missions, and aerial targeting. These flexible applications have prompted a propelled research for expanding the level of self-sufficiency of these unmanned aerial vehicles. On the premise of their configuration unmanned aerial vehicles are separated into two primary classes: fixed-wing and rotary-wing UAVs

Fixed wing UAVs is the most widely recognized sort. Their configuration is straightforward when contrasted with different sorts and they can fly for long lengths of time at high speeds. Nonetheless, these UAVs need runways or other recovery and launch frameworks like sling or parachute for takeoff and landing. Fixed wing UAVs are not suitable to be flown indoors, at low altitudes. Rotary wing UAVs, are better because of their absence of requirement for runways for takeoff and landing which makes them perfect for use in tight and congested regions. Because of their high mobility, small size and hovering capacities, rotary wing UAVs are favored.

## 2. Working Principle

A Quad Copter is a four-rotor helicopter. It is an under-actuated, dynamic vehicle with four input forces (one for each rotor) and six degrees of freedom (6DOF). Unlike regular helicopters that have variable pitch angle rotors, a quad copter has four fixed- pitch and fixed-angle rotors. The motion of a Quad copter in 6DOF is controlled by varying the rpm of the four rotors individually, thereby changing the lift & rotational forces. Quad copter tilts toward the direction of slow spinning motor, which enables it to roll and pitch. Roll and pitch angles divides the thrust into two directions due to which linear motion is achieved. The rotors rotate in clockwise-anticlockwise pairs, (figure 1) to control the yaw produced due to the drag force on propellers. The center of gravity (CG) lies almost at the same plane which contains all the rotors. Also all four motors of same class differ in efficiency with each other. This differentiates it from helicopters and it is very difficult to stabilize a quad copter by human control. Therefore a sophisticated control is essential for a balance flight of quad copter.

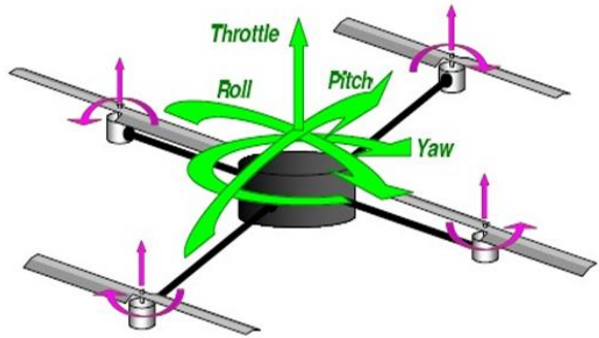

**Figure 1. Quad copter schematic**

## 3. Quad Copter Modelling

### 3.1. Quad Copter Dynamics

The motion of Quad Copter in 6DOF is controlled by varying the rpm of four rotors individually, thereby changing the vertical, horizontal and rotational forces as depicted in figure 2.

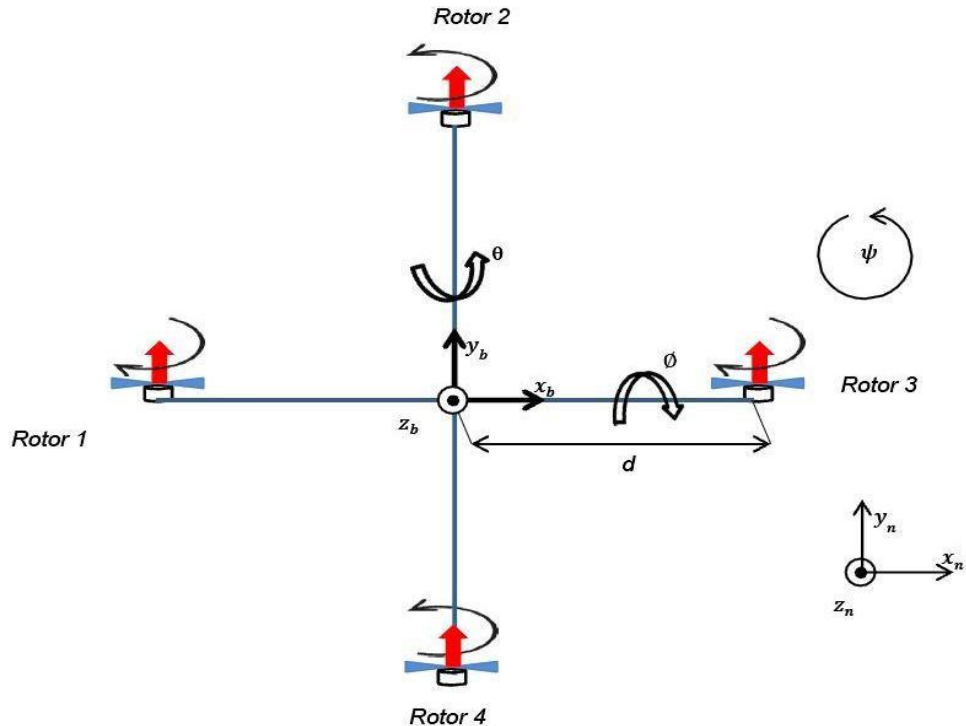

**Figure 2. Quad copter dynamics**

**3.1.1. Roll Motion:** The torque about x-axis is modeled as $\tau_x$ i.e. rolling moment. Now *r* (moment arm) is distance of individual forces from x-axis.

$$\tau_x = \sum_{i=1}^{4} r_i f_i \tag{1}$$

$$\tau_x = r_1 f_1 + r_2 f_2 + r_3 f_3 + r_4 f_4 \tag{2}$$

As the rotors 1 and 3 are placed along with body x-axis so their moment arm is zero and they do not contribute in rolling moment as shown in figure 2.

$$r_1 = r_3 = 0$$

The moment arm for rotor 2 is positive as it will create positive moment (counterclockwise), while for that of rotor 4 is negative because it will tend to rotate the system in clockwise fashion and as a convention clockwise torque is taken as negative.

$$r_2 = d$$
$$r_4 = -d$$

So equation 2 becomes:

$$\tau_x = d f_2 - d f_4 \tag{3}$$

Thus, when rotor 4 applies force it tends to rotate the system about x-axis opposite to curl of fingers (figure 2) so using the right hand rule (RHR) its torque contribution in x-axis is negative. Similarly the rotor 2 produces positive torque about x-axis according to RHR.

**3.1.2. Pitch Motion:** The torque about y-axis is modeled as $\tau_y$ i.e. pitching moment. Now $r$ (moment arm) is distance of individual forces form y-axis. Using Equation 1, we have:

$$\tau_y = r_1 f_1 + r_2 f_2 + r_3 f_3 + r_4 f_4 \tag{4}$$

Where $r$ is distance of individual force form y-axis

$$r_2 = r_4 = 0$$
$$r_1 = -r_3 = d$$

So equation 3 becomes:

$$\tau_y = d f_1 - d f_3 \tag{5}$$

Thus, when rotor 3 applies force it tends to rotate the system about y-axis opposite to curl of fingers (figure 2) so using RHR its torque contribution in y-axis is negative. Similarly the rotor 1 produces positive torque about x-axis according to RHR.

**3.1.3. Yaw Motion:** Finally the yawing moment is modeled as $\tau_z$ and is the torque about z-axis (figure 2 & 3). As the motors rotate the rotors encounter a reactive torque due to air drag on rotor blade, this reactive torque is modeled as Q. The yaw is accomplished by rotor rpm imbalance between clockwise and counter-clockwise rotating propellers. The reactive torque as a convention is taken positive for counter clockwise rotating motors/rotors and negative for clockwise rotating ones. All the four rotors contribute in yawing moment.

$$\tau_z = \sum_{i=1}^{4} Q_i \tag{6}$$

Whereas $Q = cF_i$ if rotation of the $i^{th}$ rotor is counterclockwise and $-cF_i$, if clockwise. So opening the summation becomes:

$$\tau_z = c(-F_1 + F_2 - F_3 + F_4) \qquad (7)$$

Where 'c' in Equation 7 is the force to moment scaling factor for a fixed motor-propeller set & can be calculated experimentally [1].

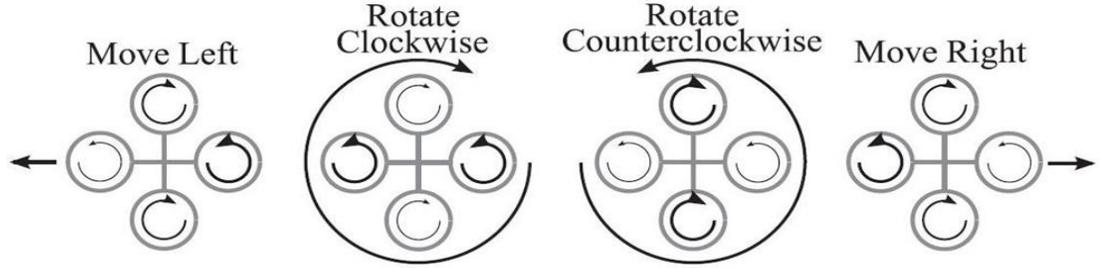

**Figure 3. Yaw & linear motion due to rotor forces**

**3.1.4. Vertical Motion:** The vertical motion of the quad copter is governed by the total thrust force 'T' and weight 'w' along the z-axis (figure 2). The rpms of the four rotating propellers are changed simultaneously in order to accomplish vertical motion.

$$T = \sum_{i=1}^{4} F_i \qquad (8)$$

The vertical motion is described by the equation (Newton's 2nd law):

$$z'' = \frac{T - mg}{m} \qquad (9)$$

**3.1.5. Horizontal Motion:** The horizontal motion in the x & y-axes is accomplished by decreasing the rpm of the rotor in whose direction it is intended to move, and subsequently increasing the rpm of the rotor on the opposite side of the same arm as depicted in figure 3. The force imbalance will cause the quad copter to tilt to one side and the horizontal component of the thrust force will impart horizontal linear motion to the quad copter (figure 4). If the mass of the quad copter is 'm', then referring to (figure 4), the linear acceleration in the horizontal x-direction can be written as:

$$x'' = -\frac{T\cos\theta}{m} \qquad (10)$$

Using small angle approximation ($\theta < 0.5$ rad):

$$\sin\theta \approx \theta \qquad (11)$$

Since $\theta$ is small:

$$T \approx mg \qquad (12)$$

Thus, from equation 10 & 11, equation 9 becomes:

$$x'' = -g\theta \tag{13}$$

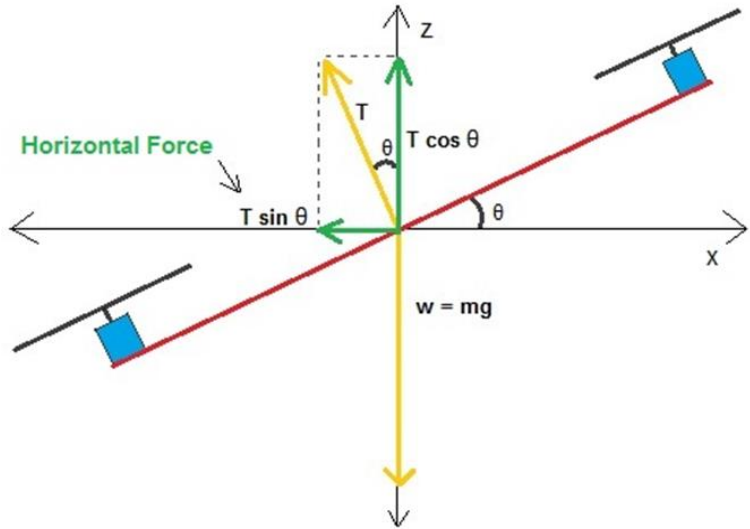

**Figure 4. Horizontal motion**

Similarly, $\ddot{y}$ can be derived out to be:

$$y'' = g\phi \tag{14}$$

Thus equation 3, 5, 7, 9, 13 & 14 give us the dynamics of the quad copter in all 6 degrees of freedom. After the dynamical equations have been established, we can move towards the state space modeling of the quad rotor. Due to the simplicity and ease of understanding, the quad rotor is first modeled in the 3DOF, then in the 6DOF setting.

### 3.2. State Space Representation

A state space representation is a mathematical model of a physical system as a set of input, output and state variables related by first-order differential equations. "State space" refers to the space whose axes are the state variables. The state of the system can be represented as a vector within that space. The most general state-space representation of a linear system with p inputs, q outputs and n state variables is written in the following form

$$\dot{x}(t) = Ax(t) + Bu(t) \tag{15}$$

$$y(t) = Cx(t) + Du(t) \tag{16}$$

Where,
$x(t)$ is called the 'State Vector'
$y(t)$ is called the 'Output Vector'
$u(t)$ is called the 'Input (or control) Vector'

A    is the 'System Matrix'
B    is the 'Input Matrix'
C    is the 'Output Matrix'
D    is the 'Feed forward Matrix'

$$\dot{x}(t) = \frac{d}{dt}x(t)$$

### 3.3. 3DOF Quad Copter Modelling

We start by selecting the system states. Intuitively, and also as seen in [2] [3] & [4], the following six states are sufficient for a quad copter in 3DOF.

    a. Roll angle - $\phi$

    b. Pitch angle - $\theta$

    c. Yaw angle - $\psi$

    d. Roll rate - $\phi'$

    e. Pitch rate - $\theta'$

    f. Yaw rate - $\psi'$

Thus, we will get our state vector $x$ as follows:

$$x^T = [\phi \; \theta \; \psi \; \phi' \; \theta' \; \psi'] \tag{17}$$

The input matrix $u$ consists of the four rotor thrusts/forces $F_i$ as in Equation 18. The numbering of rotor forces is as per (fig 2).

$$u^T = [F_1 \; F_2 \; F_3 \; F_4] \tag{18}$$

Since the quad copter is intended to be stabilized in 3DOF, the output matrix $y$ is given as follows:

$$y^T = [\phi \; \theta \; \psi] \tag{19}$$

According to equation 15, we need the state differential equations of all the states in order to get the state space model of our system. According to the Newton's 2nd law (applied to rotational motion), for any physical body with a rotational moment of inertia $I$, an angular acceleration $\varphi''$ and an applied torque $\tau$ we have the relation:

$$\tau = I\varphi'' \tag{20}$$

Thus, from equations 3, 5, 7 and 20, the state differential equations can be written as:
$\phi' = \phi'$
$\theta' = \theta'$
$\psi' = \psi'$

$$\phi'' = \frac{d}{I_x}(F_2 - F_4)$$

$$\theta'' = \frac{d}{I_y}(F_1 - F_3)$$

$$\psi'' = \frac{c}{I_z}(F_2 + F_4 - F_1 - F_3)$$

Where $I_x, I_y$ and $I_z$ are the moment of inertia of the quad copter along the x, y & z axes respectively. The state differential equations written in matrix form are shown as Equation 21 as follows. (Note that Equation 15 has been written above Equation 21 for easy comparison):

$$x' = A x + B u$$

$$\begin{bmatrix} \varphi' \\ \theta' \\ \psi' \\ \varphi'' \\ \theta'' \\ \psi'' \end{bmatrix} = \begin{bmatrix} 0 & 0 & 0 & 1 & 0 & 0 \\ 0 & 0 & 0 & 0 & 1 & 0 \\ 0 & 0 & 0 & 0 & 0 & 1 \\ 0 & 0 & 0 & 0 & 0 & 0 \\ 0 & 0 & 0 & 0 & 0 & 0 \\ 0 & 0 & 0 & 0 & 0 & 0 \end{bmatrix} \begin{bmatrix} \varphi \\ \theta \\ \psi \\ \varphi' \\ \theta' \\ \psi' \end{bmatrix} + \begin{bmatrix} 0 & 0 & 0 & 0 \\ 0 & 0 & 0 & 0 \\ 0 & 0 & 0 & 0 \\ 0 & \frac{d}{I_x} & 0 & -\frac{d}{I_x} \\ \frac{d}{I_y} & 0 & -\frac{d}{I_y} & 0 \\ -\frac{c}{I_z} & \frac{c}{I_z} & -\frac{c}{I_z} & \frac{c}{I_z} \end{bmatrix} \begin{bmatrix} F_1 \\ F_2 \\ F_3 \\ F_4 \end{bmatrix} \quad (21)$$

Comparing equation 21 with equation 15, we can observe that:

$$A = \begin{bmatrix} 0 & 0 & 0 & 1 & 0 & 0 \\ 0 & 0 & 0 & 0 & 1 & 0 \\ 0 & 0 & 0 & 0 & 0 & 1 \\ 0 & 0 & 0 & 0 & 0 & 0 \\ 0 & 0 & 0 & 0 & 0 & 0 \\ 0 & 0 & 0 & 0 & 0 & 0 \end{bmatrix} \qquad B = \begin{bmatrix} 0 & 0 & 0 & 0 \\ 0 & 0 & 0 & 0 \\ 0 & 0 & 0 & 0 \\ 0 & \frac{d}{I_x} & 0 & -\frac{d}{I_x} \\ \frac{d}{I_y} & 0 & -\frac{d}{I_y} & 0 \\ -\frac{c}{I_z} & \frac{c}{I_z} & -\frac{c}{I_z} & \frac{c}{I_z} \end{bmatrix}$$

Comparing the output equation (in matrix form) with equation 16, we have:

$$y = C x + D u$$

$$\begin{bmatrix} \varphi \\ \theta \\ \psi \end{bmatrix} = \begin{bmatrix} 1 & 0 & 0 & 0 & 0 & 0 \\ 0 & 1 & 0 & 0 & 0 & 0 \\ 0 & 0 & 1 & 0 & 0 & 0 \end{bmatrix} \begin{bmatrix} \varphi \\ \theta \\ \psi \\ \varphi' \\ \theta' \\ \psi' \end{bmatrix} + \begin{bmatrix} 0 & 0 & 0 & 0 \\ 0 & 0 & 0 & 0 \\ 0 & 0 & 0 & 0 \end{bmatrix} \begin{bmatrix} F_1 \\ F_2 \\ F_3 \\ F_4 \end{bmatrix} \quad (22)$$

Thus,

$$C = \begin{bmatrix} 1 & 0 & 0 & 0 & 0 & 0 \\ 0 & 1 & 0 & 0 & 0 & 0 \\ 0 & 0 & 1 & 0 & 0 & 0 \end{bmatrix} \qquad D = \begin{bmatrix} 0 & 0 & 0 & 0 \\ 0 & 0 & 0 & 0 \\ 0 & 0 & 0 & 0 \end{bmatrix}$$

Hence from equations 21 & 22, we have the $A, B, C$ and $D$ matrices which completely define the 3DOF state space model of a quad copter. Notice that the frictional and drag losses have been neglected.

### 3.4. 6DOF Quad Copter Modelling

We start by selecting the system states. As seen in literature [5] [6] [7] [8] and [11], the following twelve states are practically suitable for a quad copter in 6DOF.

   a. Position along x axis - $x$

   b. Position along y axis - $y$

   c. Position along z axis (height) - $z$

   d. Velocity along x axis - $x'$

   e. Velocity along y axis - $y'$

   f. Velocity along z axis - $z'$

   g. Roll angle - $\phi$

   h. Pitch angle - $\theta$

   i. Yaw angle - $\psi$

   j. Roll rate - $\phi'$

   k. Pitch rate - $\theta'$

   l. Yaw rate - $\psi'$

Thus, we will get our state vector $x$ as follows:

$$x^T = [x\ y\ z\ x'\ y'\ z'\ \phi\ \theta\ \psi\ \phi'\ \theta'\ \psi'] \tag{23}$$

The input matrix $u$ is given in Equation 21 as follows:

$$u^T = [U_1\ U_2\ U_3\ U_4] \tag{24}$$

Where,

   a. $U_1$ is the Total Upward Force on the quad rotor along z-axis ($T - mg$)

   b. $U_2$ is the Pitch Torque (about x-axis)

   c. $U_3$ is the Roll Torque (about y-axis)

   d. $U_4$ is the Yaw Torque (about z-axis)

The choice of input vector here is different than that of the 3DOF setting due to ease of modeling, as will become apparent while observing the state differential equations. The output matrix $y$ is given as follows:

$$y^T = [x \; y \; z \; \phi \; \theta \; \psi]  \quad (25)$$

From equations 3, 5, 7, 9, 13, 14 and 20, the state differential equations can be written as:

$$x' = x'$$
$$y' = y'$$
$$z' = z'$$
$$x'' = -g\theta$$
$$y'' = g\phi$$
$$z'' = -\frac{U_1}{m}$$
$$\phi' = \phi'$$
$$\theta' = \theta'$$
$$\psi' = \psi'$$
$$\phi'' = \frac{U_2}{I_x}$$
$$\theta'' = \frac{U_3}{I_y}$$
$$\psi'' = \frac{U_4}{I_z}$$

The above state differential equations written in matrix form are shown as Equation 26 as follows. Similar to the previous section (3DOF), a one to one correspondence has been done with equation 15 for the extraction of state space matrices

$$x' = A x + B u$$

$$\begin{bmatrix} x' \\ y' \\ z' \\ x'' \\ y'' \\ z'' \\ \phi' \\ \theta' \\ \psi' \\ \phi'' \\ \theta'' \\ \psi'' \end{bmatrix} = \begin{bmatrix} 0 & 0 & 0 & 1 & 0 & 0 & 0 & 0 & 0 & 0 & 0 & 0 \\ 0 & 0 & 0 & 0 & 1 & 0 & 0 & 0 & 0 & 0 & 0 & 0 \\ 0 & 0 & 0 & 0 & 0 & 1 & 0 & 0 & 0 & 0 & 0 & 0 \\ 0 & 0 & 0 & 0 & 0 & 0 & 0 & -g & 0 & 0 & 0 & 0 \\ 0 & 0 & 0 & 0 & 0 & 0 & g & 0 & 0 & 0 & 0 & 0 \\ 0 & 0 & 0 & 0 & 0 & 0 & 0 & 0 & 0 & 0 & 0 & 0 \\ 0 & 0 & 0 & 0 & 0 & 0 & 0 & 0 & 0 & 1 & 0 & 0 \\ 0 & 0 & 0 & 0 & 0 & 0 & 0 & 0 & 0 & 0 & 1 & 0 \\ 0 & 0 & 0 & 0 & 0 & 0 & 0 & 0 & 0 & 0 & 0 & 1 \\ 0 & 0 & 0 & 0 & 0 & 0 & 0 & 0 & 0 & 0 & 0 & 0 \\ 0 & 0 & 0 & 0 & 0 & 0 & 0 & 0 & 0 & 0 & 0 & 0 \\ 0 & 0 & 0 & 0 & 0 & 0 & 0 & 0 & 0 & 0 & 0 & 0 \end{bmatrix} \begin{bmatrix} x \\ y \\ z \\ x' \\ y' \\ z' \\ \phi \\ \theta \\ \psi \\ \phi' \\ \theta' \\ \psi' \end{bmatrix} + \begin{bmatrix} 0 & 0 & 0 & 0 \\ 0 & 0 & 0 & 0 \\ 0 & 0 & 0 & 0 \\ 0 & 0 & 0 & 0 \\ 0 & 0 & 0 & 0 \\ \frac{1}{m} & 0 & 0 & 0 \\ 0 & 0 & 0 & 0 \\ 0 & 0 & 0 & 0 \\ 0 & 0 & 0 & 0 \\ 0 & \frac{1}{I_x} & 0 & 0 \\ 0 & 0 & \frac{1}{I_y} & 0 \\ 0 & 0 & 0 & \frac{1}{I_z} \end{bmatrix} \begin{bmatrix} U_1 \\ U_2 \\ U_3 \\ U_4 \end{bmatrix} \quad (26)$$

Comparing equation 26 with equation 15, we can observe that:

$$A = \begin{bmatrix} 0 & 0 & 0 & 1 & 0 & 0 & 0 & 0 & 0 & 0 & 0 & 0 \\ 0 & 0 & 0 & 0 & 1 & 0 & 0 & 0 & 0 & 0 & 0 & 0 \\ 0 & 0 & 0 & 0 & 0 & 1 & 0 & 0 & 0 & 0 & 0 & 0 \\ 0 & 0 & 0 & 0 & 0 & 0 & 0 & -g & 0 & 0 & 0 & 0 \\ 0 & 0 & 0 & 0 & 0 & 0 & g & 0 & 0 & 0 & 0 & 0 \\ 0 & 0 & 0 & 0 & 0 & 0 & 0 & 0 & 0 & 0 & 0 & 0 \\ 0 & 0 & 0 & 0 & 0 & 0 & 0 & 0 & 1 & 0 & 0 & 0 \\ 0 & 0 & 0 & 0 & 0 & 0 & 0 & 0 & 0 & 1 & 0 & 0 \\ 0 & 0 & 0 & 0 & 0 & 0 & 0 & 0 & 0 & 0 & 1 & 0 \\ 0 & 0 & 0 & 0 & 0 & 0 & 0 & 0 & 0 & 0 & 0 & 0 \\ 0 & 0 & 0 & 0 & 0 & 0 & 0 & 0 & 0 & 0 & 0 & 0 \\ 0 & 0 & 0 & 0 & 0 & 0 & 0 & 0 & 0 & 0 & 0 & 0 \end{bmatrix} \qquad B = \begin{bmatrix} 0 & 0 & 0 & 0 \\ 0 & 0 & 0 & 0 \\ 0 & 0 & 0 & 0 \\ 0 & 0 & 0 & 0 \\ \frac{1}{m} & 0 & 0 & 0 \\ 0 & 0 & 0 & 0 \\ 0 & 0 & 0 & 0 \\ 0 & 0 & 0 & 0 \\ 0 & 0 & 0 & 0 \\ 0 & \frac{1}{I_x} & 0 & 0 \\ 0 & 0 & \frac{1}{I_y} & 0 \\ 0 & 0 & 0 & \frac{1}{I_z} \end{bmatrix}$$

Comparing the output equation 27 (in matrix form) with equation 16, we have:

$$y = C x + D u$$

$$\begin{bmatrix} x \\ y \\ z \\ \varphi \\ \theta \\ \psi \end{bmatrix} = \begin{bmatrix} 1 & 0 & 0 & 0 & 0 & 0 & 0 & 0 & 0 & 0 & 0 & 0 \\ 0 & 1 & 0 & 0 & 0 & 0 & 0 & 0 & 0 & 0 & 0 & 0 \\ 0 & 0 & 1 & 0 & 0 & 0 & 0 & 0 & 0 & 0 & 0 & 0 \\ 0 & 0 & 0 & 0 & 0 & 0 & 1 & 0 & 0 & 0 & 0 & 0 \\ 0 & 0 & 0 & 0 & 0 & 0 & 0 & 1 & 0 & 0 & 0 & 0 \\ 0 & 0 & 0 & 0 & 0 & 0 & 0 & 0 & 1 & 0 & 0 & 0 \end{bmatrix} \begin{bmatrix} x \\ y \\ z \\ x' \\ y' \\ z' \\ \varphi \\ \theta \\ \psi \\ \varphi' \\ \theta' \\ \psi' \end{bmatrix} + \begin{bmatrix} 0 & 0 & 0 & 0 \\ 0 & 0 & 0 & 0 \\ 0 & 0 & 0 & 0 \\ 0 & 0 & 0 & 0 \\ 0 & 0 & 0 & 0 \\ 0 & 0 & 0 & 0 \end{bmatrix} \begin{bmatrix} U_1 \\ U_2 \\ U_3 \\ U_4 \end{bmatrix} \qquad (27)$$

Thus,

$$C = \begin{bmatrix} 1 & 0 & 0 & 0 & 0 & 0 & 0 & 0 & 0 & 0 & 0 & 0 \\ 0 & 1 & 0 & 0 & 0 & 0 & 0 & 0 & 0 & 0 & 0 & 0 \\ 0 & 0 & 1 & 0 & 0 & 0 & 0 & 0 & 0 & 0 & 0 & 0 \\ 0 & 0 & 0 & 0 & 0 & 0 & 1 & 0 & 0 & 0 & 0 & 0 \\ 0 & 0 & 0 & 0 & 0 & 0 & 0 & 1 & 0 & 0 & 0 & 0 \\ 0 & 0 & 0 & 0 & 0 & 0 & 0 & 0 & 1 & 0 & 0 & 0 \end{bmatrix} \qquad D = \begin{bmatrix} 0 & 0 & 0 & 0 \\ 0 & 0 & 0 & 0 \\ 0 & 0 & 0 & 0 \\ 0 & 0 & 0 & 0 \\ 0 & 0 & 0 & 0 \\ 0 & 0 & 0 & 0 \end{bmatrix}$$

From equations 26 & 27, we have the $A, B, C$ and $D$ matrices which completely define the 6DOF state space model of a quad copter.

## 4. Conclusion

3DOF and 6DOF state space models of the quad copter have been derived from basic Newtonian equations in this paper. The quad copter mathematical model derived in this paper is unique in the sense that it gives the reader the grass root level understanding of the dynamics of the quad copter UAV in 3DOF and 6DOF dynamical system. The models derived here will be later used for designing control laws for the 3DOF and 6DOF stability of the quad copter.